\documentclass[letterpaper, 10 pt, conference]{ieeeconf}  % Comment
\usepackage{tikz}
\usetikzlibrary{arrows.meta}
\usetikzlibrary{decorations.pathreplacing}
\usetikzlibrary{calc,patterns,angles,quotes}
\usetikzlibrary{shapes.misc}

\usepackage{amsfonts}
\usepackage{amsmath}
\usepackage{mathtools}
\usepackage{multicol}
\usepackage{subcaption}
\usepackage{pgffor}
\usepackage{pgfplots}
\usepackage{tikz-3dplot}
\usepackage{xcolor}
\usepackage{float}
\usepackage{hyperref} % For clicable references
\usepackage[noabbrev]{cleveref}
\usepackage{bm}
\usepackage{algorithm}
\usepackage[noend]{algpseudocode}
\usepackage{array}
\usepackage{csvsimple}
\usepackage{readarray}
\usepackage{datatool}

\pgfplotsset{compat=1.8}
\usetikzlibrary{shapes.misc}
\usetikzlibrary{calc}
\readarraysepchar{,}
\crefformat{equation}{(#2#1#3)}
\crefrangeformat{equation}{(#3#1#4) to~(#5#2#6)}
\crefmultiformat{equation}{(#2#1#3)}%
{ and~(#2#1#3)}{, (#2#1#3)}{ and~(#2#1#3)}

\usepackage{amssymb}% http://ctan.org/pkg/amssymb
\usepackage{pifont}% http://ctan.org/pkg/pifont

\usepackage{booktabs}% http://ctan.org/pkg/booktabs
\usepackage{multirow}% http://ctan.org/pkg/booktabs
\hypersetup{
    colorlinks,
    linkcolor={red!50!black},
    citecolor={blue!50!black},
    urlcolor={blue!80!black}
  }
  
\IEEEoverridecommandlockouts
\newtheorem{remark}{Remark}
\newtheorem{theorem}{Theorem}
\newtheorem{proposition}{Proposition}

\newtheorem{example}{Example}

\newcommand{\ix}{{\pmb{x}}}
\newcommand{\iy}{{\pmb{y}}}
\newcommand{\iu}{{\pmb{u}}}
\newcommand{\iv}{{\pmb{v}}}
\newcommand{\iz}{{\pmb{z}}}

\newcommand{\ic}{{\pmb{c}}}
\newcommand{\ip}{{\pmb{p}}}
\newcommand{\iq}{{\pmb{q}}}

\newcommand{\ialpha}{{\pmb{\alpha}}}
\newcommand{\ibeta}{{\pmb{\beta}}}

\newcommand{\R}{{\mathbb R}}
\newcommand{\N}{{\mathbb N}}
\newcommand{\M}{{\mathcal M}}
\newcommand{\D}{{\mathcal D}}
\newcommand\PM[2]{\M_{#1,#2}}

\newcommand{\xinit}{{\ix_0}}
\newcommand{\xfinal}{{\ix_T}}

\newcommand{\configspace}{{\mathcal X}}

\usepackage{titlesec}
\titlespacing*{\section}{2pt}{0.1\baselineskip}{0.1\baselineskip}
\titlespacing*{\subsection}{2pt}{0.\baselineskip}{0\baselineskip}

\title{\bf  Piecewise-Linear Motion Planning\\ amidst Static, Moving, or Morphing Obstacles}

\author{Bachir El Khadir$^{1}$, Jean Bernard Lasserre$^{2}$,  Vikas Sindhwani$^{3}$% <-this % stops a space
\thanks{$^{1}$ IBM Watson Research Center, Yorktown Heights, NY 10598, USA {\tt\small bachir@ibm.com}}% <-this % stops a space
\thanks{$^{2}$ Laboratoire d’Analyse et d'Architecture des Syst\`emes (LAAS), Institute of Mathematics, University of Toulouse, France {\tt\small lasserre@laas.fr}
 J.B. Lasserre was partly funded by the AI Interdisciplinary Institute ANITI through the French ``Investing for the Future PI3A" program under the Grant agreement ANR-19-PI3A-0004
}%
\thanks{$^{3}$ Robotics at Google, New York City, NY 10011, USA
        {\tt\small sindhwani@google.com}}%
}

\begin{document}
\maketitle

% The following two lines generate page numbers
\thispagestyle{plain} 
\pagestyle{plain}

\begin{abstract}
  We propose a novel method for planning shortest length piecewise-linear
  motions through complex environments punctured with static, moving, or even
  morphing obstacles. Using a moment optimization approach, we formulate a
  hierarchy of semidefinite programs that yield increasingly refined lower
  bounds converging monotonically to the optimal path length.
  For computational tractability,
  our global moment optimization approach motivates an iterative motion planner
  that outperforms competing sampling-based and nonlinear optimization
  baselines. Our method natively handles continuous time constraints without any
  need for time discretization, and has the potential to scale better with
  dimensions compared to popular sampling-based methods.
\end{abstract}

\section{Introduction and Problem Statement}

How should robots -- viewed as complex systems of articulated rigid
bodies -- move from a start to a goal configuration in an environment
cluttered with static and dynamic obstacles? Even without considering
dynamic feasibility of a desired motion, mechanical and sensor
limitations, uncertainty and feedback, the purely geometric motion
planning problem is known to be computationally
hard~\cite{reif1979complexity} in its full generality.

%A fundamental problem in robotics is to plan the trajectory of a robot
%in an environment filled with moving obstacles.

% \vikas{Can the basic problem below be written in C-space formalism?}
% \vikas{A picture: moving obstacles described by polynomial inequalities, and solutions at various levels, would be useful here as a preview of the salient points of the paper}

%% \subsection{Problem Statement}
\subsection{The Optimal Motion Planning Problem}
We follow a similar notation to that of \cite{karaman2011sampling} to describe the Optimal Motion Planning (OMP) problem. Let $\configspace = \R^{n}$ be the configuration space, where $n \in \N$.
We are interested in finding the shortest path
$\ix: [0, T] \rightarrow \configspace$ (where $T$ is a positive constant) that starts
at a configuration $\ix(0) = \xinit \in \configspace$, ends at a configuration
$\ix(T) = \xfinal \in \configspace$, and avoids  a time-varying obstacle region
$\configspace_{{\text{obs}}}(t) \subseteq \configspace$ at all times $t \in [0, T]$. Here,
we assume that the obstacle-free space $\configspace_{\text{free}}(t) \coloneqq \configspace \setminus \configspace_{{\text{obs}}}(t)$ is a \emph{closed basic semialgebraic} set, i.e., that there exists a (multivariate, scalar-valued)
polynomial function $g_k \in \mathbb R[t, \ix]$ in variables $t$ and $\ix$ such that
\[\configspace_{\text{free}}(t) \coloneqq \{ \ix \in \mathbb R^n \; | \; g_1(t, \ix) \ge 0, \ldots, g_{k}(t, \ix) \ge 0\}.\]

\begin{figure}%[ht]
\resizebox{250pt}{!}{%
\begin{tikzpicture}
%\draw[help lines, color=gray!30, dashed] (0,-.9) grid (.9,.9);
\foreach \i in {0,2,4,6,8,10} {
    \coordinate (A\i) at ($(\i,0)+(0.5cm,0)$) {};

    \draw ($(A\i)+(0,20pt)$) node[inner sep=0] {{\includegraphics[trim={13cm 12cm 13cm 13cm},clip,scale=.15]{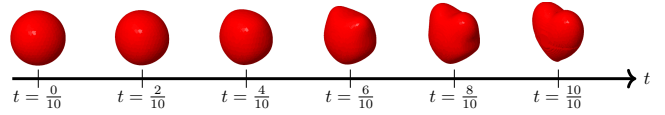}}};
    \draw ($(A\i)+(0,5pt)$) -- ($(A\i)-(0,5pt)$);
    \draw ($(A\i)-(0,10pt)$) node {$t = \frac{\i}{10}$};
}
\draw[->,ultra thick] (0,0)--(12,0) node[right]{$t$};

\end{tikzpicture}
}
\caption{\label{fig:sphere_to_heart} \small Example of an time-varying obstacle
  described by the polynomial inequality $g(t, \ix) < 0$, with
  $g(t, \ix) \coloneqq (1-t)(x_1^2 + x_2^2 + x_3^2+1) + \frac t {320} (320x_1^2x_3^3 + 36x_2^2x_3^3 - 5(4x_1^2 + 9x_2^2 + 4x_3^2 - 4)^3)$.
  The shape of the obstacle changes from a sphere to a heart as time $t$ goes
  from $0$ to $1$.}
\vspace{-.7cm}
\end{figure}

Our choice for working with polynomial functions to describe obstacles stems
from two reasons. On the one hand, polynomial functions can uniformly
approximate any continuous function over compact sets, and hence are powerful
enough for modeling purposes. See \cref{fig:sphere_to_heart} for an illustration
of an obstacle morphing into a complex shape over time, as described by a
degree-$7$ polynomial, and see
\cite{ahmadi2016geometry,dabbene2013set,dabbene2017simple,pauwels2016sorting}
for more examples on the use of polynomial functions for the purposes of
modeling 3D geometry. On the other hand, as we will see in
\cref{sec:background-sos}, the discovery of recent connections between algebraic
geometry and semidefinite programming has resulted in powerful tools that are
designed specifically for tackling optimization problems which are described by
polynomial data.

More formally, the OMP problem described by data
\begin{equation}\D = (\xinit, \xfinal, \{g_1, \ldots, g_m\})\label{eq:data_D},
\end{equation}
where $\xinit, \xfinal \in \R^{n}$ and $g_1, \ldots, g_{m} \in \R[t, \ix]$ is the following minimization problem,
\begin{equation}
  \hspace{-.3cm}
   \boxed{
  \tag*{OMP(\(\D\))}
  \label{eq:opt_shortest_trajectory}
  \begin{split}
  &\min_{\ix: [0, T] \rightarrow \R^{n}} \quad  \int_0^T \| \dot \ix(t) \| \; {\rm d}t\\
  &\;\textrm{s.t.}  \quad \ix(0) = \xinit \,,\: \ix(T) = \xfinal,\\
  &\;\quad\quad g_k(t, \ix(t)) \ge 0 \quad \forall t \in [0, T],\;\forall k \in [m],
  \end{split}
  }
\end{equation}
where $\|\cdot\|$ denotes the $\ell_2$ norm, and $[s]$ denotes the set
$\{1, \ldots, s\}$.  The objective term
$\int_0^T \| \dot \ix(t) \| \; {\rm d}t$ is the length of the
path $\ix(t)$.
A path that satisfies the constraints of
\ref{eq:opt_shortest_trajectory} is said to be \emph{feasible}. A path that is feasible and has minimum length is said to be \emph{optimal}.

%%% Local Variables:
%%% mode: latex
%%% TeX-master: "../Path_Planning_Using_Moments"
%%% End:

\subsection{Background on Motion Planning}

We first set the stage for describing and motivating our approach in relation to
the vast prior
literature~\cite{lavalle1999planning,halperin2017algorithmic,lynch2017modern} on
motion planning. Most obviously, \ref{eq:opt_shortest_trajectory} can be
transcribed into a nonlinear optimization problem by using a parametric
representation of the path together with time-discretization to construct a
finite-dimensional optimization
problem~\cite{schulman2014motion,kalakrishnan2011stomp,sindhwani2017sequential}.
Because of its non-convexity, the effectiveness of such an approach depends on
having a good initial guess and in general no guarantees can be provided that
the process will not return a sub-optimal stationary point. Closely related is
the body of work on virtual potential fields~\cite{khatib1986real} where a
vector field is designed to pull the robot towards the goal and push it away
from obstacles. Unless a restricted class of {\it navigation
  functions}~\cite{rimon1992exact} generates the gradient flow, these methods
are also susceptible to local minima. By contrast, sampling-based motion
planners~\cite{lavalle2001rapidly,karaman2011sampling,janson2018deterministic},
pervasively used in robotics, are attractive since they can at-least offer a
guarantee of probabilistic completeness, that is, as the planning time goes to
infinity, the probability of finding a solution tends to one. Sampling-based
planners rely on a collision checking primitive to construct a data structure,
e.g., tree or a graph, that stores a sampling of obstacle-avoiding feasible
motions of the robot. In their most common instantiations, sampling methods
return feasible paths, not necessarily optimal~\cite{karaman2011sampling}, almost alway
also requiring post processing to reduce jerkiness.

Even in the time-independent, purely geometric path planning setting, the general problem of finding a feasible path, or correctly reporting that such a path does not exist, has been shown by Reif~\cite{reif1979complexity} to be PSPACE-hard. If ${\cal X }_{\text{free}}$ is semialgebraic, then its cylindrical cell decomposition~\cite{schwartz1984piano} allows for a doubly-exponential (in the configuration space dimension $n$) solution to the motion planning problem. Canny's Roadmap~\cite{canny1988complexity} gives an improved single-exponential solution based on the notion of a {\it roadmap}, a network of one-dimensional curves preserving the connectivity of the free space that can be reached from any configuration. However, despite their completeness guarantee, these techniques are considered computationally impractical for all but simple or low-degree-of-freedom problems.
%\vikas{Complete algorithms, complexity, time-dependent problems}.

It should be no surprise that dynamic environments where obstacles can appear, disappear, move or morph only magnify the hardness of general motion planning~\cite{reif1994motion}, even when the obstacle motion is pre-specifed as a function of time. Many planners can be adapted to this setting by simply defining the problem in a time-augmented state space. Then, the primary complication stems from the requirement that time must always increase along a path. An alternative is to decouple space and time planning by first finding a collision-free path in the absence of moving obstacles, and then determining a time scaling function. In any case, planners for time-varying problems may also become prone to failure simply due to discretization of time.

\subsection{Statement of Contributions}

In this paper, we focus on solving the optimal motion planning problem for
piecewise-linear motions. At the outset, it should be noted that even with this
restriction, the problem remains PSPACE-hard~\cite{sutner1988motion}. With this
setting, our contributions are as follows. First, we introduce a new arsenal of
algorithmic and complexity-characterization tools from polynomial optimization
and semidefinite programming (SDP) to the motion planning literature.
Specifically, for any optimal motion problem \ref{eq:opt_shortest_trajectory}
described by data $\D$ as in \eqref{eq:data_D}, and for any number of pieces
$s$, we present a hierarchy of semidefinite programs \ref{eq:sdp_pl_trajectory}
indexed by a scalar~$r$. Every level of this hierarchy provides a lower bound
$\rho(r, s; \D)$ on the minimum length $\rho(s; \D)$ attained by
piecewise-linear paths that are feasible to \ref{eq:opt_shortest_trajectory} and
have $s$ pieces.
Importantly, we provide the asymptotic guarantee that $\rho(r, s; \D) \rightarrow\rho(s; \D)$ as $r \rightarrow \infty$. This notion of asymptotic completeness is analogous to probabilistic completeness in sampling-based methods, in the sense that in the limit of increasing computation, we are guaranteed to optimally solve the problem, or declare that no solution exists.

To remain computationally competitive with practical motion planners, we also
derive a sequential SDP-based method called \texttt{Moment Motion Planner}
(MMP). Unlike previously proposed planners for dynamic obstacle avoidance, MMP
natively handles continuous-time constraints, does not require any
discretization, and relies on semidefinite programs whose size scales polynomially
in configuration space dimensionality. On several benchmark problems involving
static, moving and morphing obstacles in dimension $2$, $3$, and $4$, including
a bimanual planar manipulation task, MMP consistently outperforms RRT and
nonlinear programming based baselines, while returning smoother paths in
comparable solve time.

\section{Moment-based approach for time-varying optimization problems}
\label{sec:background-sos}

% The moment approach to obtain convex relaxations of optimization
% problems that are described by polynomial functions.
The last few decades have known the emergence of a powerful \emph{moment-based
  approach} for solving optimizaton problems that are described by polynomial
data~\cite{lasserrecup2015}. One of the main challenges one faces when applying
this moment approach to the motion planning problem
\ref{eq:opt_shortest_trajectory} is the fact that solutions (and the constraints
on these solutions) vary continuously with time. For clarity of presentation, we
first ignore the complexities arising from this time dependence and present the
basic ideas behind this approach. Then, we present a result from real algebraic
geometry on sum of squares representations of univariate polynomial matrices
that will allow us to impose time-varying constraints on time-varying solutions.

Let us recall some standard notation. For any vector
$\ialpha \in \N^n$, $|\ialpha|$ denotes $\sum_{i=1}^n \alpha_i$. We
denote by $\N^n_d$ the set of vectors $\ialpha \in \N^n$ that satisfy
$|\ialpha| \le d$.  We denote by $\R[\iy]$ the set of (scalar valued)
polynomial functions in the variables $y_1, \ldots, y_n$. For
$\ialpha \in \N^n$, the monomial
$y_1^{\alpha_{1}} \ldots y_n^{\alpha_{n}}$ is denote by
$\iy^\ialpha$, and the coefficient of a polynomial $p \in \R[\iy]$
corresponding to the monomial $\iy^\ialpha$ is denoted by
$p_\ialpha$. 
The degree of the monomial $\iy^\ialpha$ is
$|\ialpha|$, and the degree $\deg p$ of a polynomial $p \in \mathbb R[\iy]$ is the
maximum degree of its monomials. We denote by $\R_d[\iy]$ the set of
polynomials of degree smaller than or equal to $d$.

\subsection{Moment Approach for Polynomial Optimization Problems}

A polynomial optimization problem is a problem of the form
\begin{equation}\tag{P}
  \begin{split}
    p^*\,=\,\min_{\iy \in \mathbb R^n}\quad& p(\iy)\\
    \text{s.t.} \quad &h_k(\iy) = 0 \quad k \in [m_1],\\
    \text{s.t.} \quad &g_k(\iy) \ge 0 \quad k \in [m_2],
\end{split}
\label{eq:poly_opt}
\end{equation}
where $p, h_1, \ldots, h_{m_1}, g_1, \ldots, g_{m_2} \in \R[\iy]$.
In general, problem \cref{eq:poly_opt} is nonconvex and is very
challenging to solve. In fact it is NP-hard even when $m_1 = 0$, $m_2
= 0$, and $p$ is a polynomial of degree four (see, e.g.,
\cite{murty1985some}).
An approach pioneered in \cite{lasserre2001global} has been to replace
the feasible set
\[K \coloneqq \{\iy \in \R^{n} \; | \; h_k(\iy) = 0 \; \forall k \in [m_1], \; g_k(\iy)  \ge 0 \; \forall k \in [m_2]\}\] with the set $\M(K)_+$ of
nonnegative Borel measures on $K$ of total mass equal to one, leading to
the optimization problem
\begin{equation}
    \min_{\mu \in \M(K)_+}\quad \int p(\iy) \; {\rm d}\mu.
\label{eq:inf_moment_opt}
\end{equation}
% It is not hard to see that the optimal value of \cref{eq:poly_opt} is
% equal to that of problem \cref{eq:inf_moment_opt}. Furthermore,
% problem \cref{eq:inf_moment_opt} has a linear objective function and
% a convex feasible set,
It is not hard to see that the optimal value of problem~\cref{eq:inf_moment_opt} is equal to that of \cref{eq:poly_opt}. Moreover,
problem \cref{eq:inf_moment_opt} has a linear objective function and a convex (infinite-dimensional)
feasible set.

We will now explain how to obtain a finite dimensional, convex relaxation of
\cref{eq:inf_moment_opt}. The key idea is to view \cref{eq:inf_moment_opt} not
as an optimization problem over measures $\mu \in \M(K)_{+}$, but as an
optimization problem over sequences of \emph{moments}
\(\{\int \iy^{\ialpha} d{\mu}\}_{\ialpha \in \N^n}\)
of measures $\mu \in \M(K)_{+}$. This is possible because the
objective function
\(\int p(\iy) {\rm d}\mu =  \sum_{|\ialpha| \le \deg p} \ip_{\ialpha}\int \iy^{\ialpha} d{\mu}\)
of problem \cref{eq:inf_moment_opt}
only depend on the measure $\mu$ through its first few moments.

%
%$p, g_1, \ldots, g_{m_1}, h_1, \ldots, h_{m_1}$.
Before we move further with the explanation of the moment approach, we
need to introduce some additional notation. For any integer $r \in \N$, we denote
by $\PM{r}{n}$ the set of truncated sequences of ``pseudo-moments'' in
$n$ variables, i.e., elements of the form
\((\phi_\ialpha)_{\ialpha \in \N_r^n},\) where $\phi_\ialpha \in \R$
for every $\ialpha \in \N_r^n$. Note that any measure $\mu$ gives rise
to an element of $\PM{r}{n}$, namely,
\( \left(\int \iy^{\ialpha} d{\mu}\right)_{ \ialpha \in \N^n_r} \in
\PM{r}{n},\)
but a general element of $\PM{r}{n}$ might not come from a
measure.
For any $\phi \in \PM{r}{n}$, we introduce the so-called \emph{Riesz functional}
$L_\phi : \R_r[\iy] \rightarrow \R$ defined by
\[q \; \left(= \sum_{\ialpha \in \N^n_r} q_\ialpha x^\ialpha\right) \mapsto \sum_{\ialpha \in \N^n_r} \phi_\ialpha q_\ialpha.\]
The functional $L_{\phi}$ is to ``pseudo-moments'' what the expectation operator
is to genuine moments. For $\phi \in \PM rn$ and $q \in \R_r[\iy]$, we denote
by $M_\phi(q)$ the \emph{localization matrix} associated with $q$ and $\phi$,
i.e., the matrix
\[M_\phi(q)_{\ialpha, \ibeta} = L_{\phi}(\iy^\ialpha \iy^{\ibeta} q(\iy)) \quad \forall \ialpha,\ibeta \in \N_{\lfloor (r-\deg q)/2 \rfloor}^n,\]
whose rows and columns are labeled by elements of
$\N_{\lfloor (r-\deg q)/2 \rfloor}^n$, where $\lfloor \cdot \rfloor$ is the floor function.
%
 %
% (We recall
% that the support of a measure $\mu \in \mathcal M$, denoted
% ${\rm supp}(\mu)$, is the smallest closed set $A \subseteq \R^{n}$
% such that $\mu(\R^{n}\setminus A)=0$.)

Now, for an integer $r$ larger than the maximum of the degrees of the
polynomials $p, h_1, \ldots, h_{m_1}, g_1, \ldots, g_{m_1}$, consider
the \emph{moment relaxation of order $r$} of problem
\cref{eq:inf_moment_opt} given by
\begin{equation}%\tag{SDPP(r)}
  \begin{aligned}
    \min_{\phi \in \PM{r}{n}}& L_{\phi}(p)\\
    \text{s.t.} \quad & L_\phi(1)\,=\,1\,;\:M_{\phi}(1) \succeq 0\\
    &L_{\phi}(\iy^{\alpha}h_k) =0 \; \forall \ialpha \in
    \N^n_{r-d_k}, \; \forall k \in [m_1],\\
    &M_{\phi}(g_k) \succeq 0, \; \forall k \in [m_2].
\end{aligned}
\label{eq:inf_moment_opt_r}
\end{equation}

To see that problem \eqref{eq:inf_moment_opt_r} is indeed a relaxation of problem~\eqref{eq:inf_moment_opt}, take an arbitrary candidate measure $\mu \in \M(K)_+$ with corresponding objective value $v \coloneqq \int p(\iy) {\rm d}\mu$ for problem \eqref{eq:inf_moment_opt}, and let us extract from it
the truncated sequence of moments
\(\phi \coloneqq (\int \iy^\ialpha {\rm d}\mu)_{|\ialpha| \le
  r} \in \PM{r}{n}\)
and show that $\phi$ is (i) feasible to problem~\eqref{eq:inf_moment_opt_r} and (ii) has $v$ as objective value.
To show (i), note that $L_\phi(1) = \int {\rm d} \mu = 1$, and that the matrix
$M_{\phi}(1)$ is positive semidefinite because for all polynomials
$q \in \R_{r}[\iy]$,
\(\iq^T M_{\phi}(1)\iq = \int q^2(\iy) \; {\rm d}\mu \ge 0,\)
where $\iq$ is the vector of coefficients of the
polynomial $q$. A similar reasoning shows that $\phi$ satisfies all
of the remaining constraints of problem \cref{eq:inf_moment_opt_r}. 
To show (ii), simply observe that
\(L_\phi(p) = \int p(\iy) {\rm d} \mu = v.\)
%
%\vikas{Doesnt the above line require that $\phi$ be the full sequence of moments, without any $r$-truncation? {\bf Bachir:} The above line should hold as soon as $r \ge \deg(p)$. I will make that point clearer in the writing.}
%om
The constraints \(L_\phi(1)\,=\,1\) and \(M_{\phi}(1) \succeq 0\) do
not depend on the data of the problem at hand. We refer to them as
\emph{moment consistency constraints}. 
%\vikas{let us add a remark on how to extract the solution to the original problem after solving this relaxation.}

In general, it is not always possible to extract an optimal solution $\iy \in \R^{n}$ of
\eqref{eq:poly_opt} from a "pseudo-moment" solution $\phi\in\mathcal{M}_{r,n}$
of \eqref{eq:inf_moment_opt_r}.
  However, under some conditons that hold generically (see, e.g.,
  \cite{Nie2013,Nie2014}), there exists an order $r$ for which the optimal value
  of \eqref{eq:inf_moment_opt_r} is equal to that of \eqref{eq:poly_opt}, and an
  optimal solution $\iy$ of \eqref{eq:poly_opt} can be recovered from $\phi$ by
  a linear algebra routine.
  % Even more, if \eqref{eq:poly_opt} has a unique
  % solution then ${\rm rank} \,M_{\phi}=1$ and one reads $\iy$ from the
  % first-order moments of $\phi$.
  For more details related to extraction of solutions from moment relaxations,
  the interested reader is referred to \cite{lasserrecup2015}.

  For any $r \in \N$, problem \cref{eq:inf_moment_opt_r} is an SDP that can be
  readily solved by off-the-shelf solvers such that MOSEK \cite{mosek}. We
  remind the reader that an SDP is the problem of optimizing a linear function
  subject to linear matrix inequalities. SDPs can be solved to arbitrary
  accuracy in polynomial time. See \cite{survey_sdp_vanderberghe} for a survey
  of the theory and applications of this subject.

% This is only needed for the proof, maybe move it there?
% As $r \rightarrow \infty$, under
% some assumptions, the optimal value of \cref{eq:inf_moment_opt_r}
% converges to that of \cref{eq:inf_moment_opt}. The following theorem
% will be useful to us to prove such convergence results.
% %
% \begin{theorem}[\cite{Lasserre_tams_2013}]
% \label{th-positivity}
%   For any measure $\mu \in \PM rn$ and any polynomial
%   $p \in \R[\iy]$, if $\supp(\mu)$ is compact and  $M_{d,\mu}(p)\succeq0$ for all
%   $d \in \N$, then $p(\iy) \geq 0$ for all $\iy \in \supp(\mu)$.
% \end{theorem}

\begin{remark}[Notation for vector-valued variables]
  In the rest of the paper, we will often deal with variables that are
  vector valued.  To lighten our notation, we use $\R[\iy_1, \ldots, \iy_s]$
  (resp. $\PM r {n_{1}+\ldots+ n_{s}}$) to denote the set of
  polynomials (resp. truncated sequences of pseudo-moments) in all of the
  entries of the vector-valued variables $\iy_1 \in \R^{n_{1}}, \ldots, \iy_s \in \R^{n_{s}}$. We
  also write $(\iy_1, \ldots, \iy_s)^{(\ialpha_1, \ldots, \ialpha_s)}$
  to denote the monomial
  $\iy_1^{\ialpha_1} \ldots \iy_s^{\ialpha_s}$, where for each
  $i \in [s]$, $\ialpha_i$ is
  an integer vector of the same size as $\iy_i$.
  \end{remark}

\subsection{Extension to the time-varying setting.}

% \vikas{Time-dependence is central to the paper. So suggest making this its own section.}

In this paper, we are interested in a variation of problem \eqref{eq:poly_opt}
where the inequality constraints are time-varying, i.e., a variation where
inequalties are of the form
\[g(t, \iy) \ge 0 \quad \forall t \in [0, T],\]
where $g \in \R[t, \iy]$. Such a constraint can be viewed as a continuum of
constraints \(g_t(\iy) \ge 0\) indexed by $t \in [0, T]$, where
$g_t \coloneqq g(t, \cdot) \in \R[\iy]$. If we denote the univariate polynomial matrix $t \mapsto M_{\phi}(g_t)$ by $X(t)$, then the moment
approach explained above leads to the constraint
\begin{equation}
X(t) \succeq  0 \quad \forall t \in [0, T].\label{eq:Xt_psd}
\end{equation}
The observation that the coefficients of the polynomial matrix $X$ depend linearly on
the elements of $\phi$ combined with~\cref{prop-matrix-positivstellensatz}  allows us to rewrite
constraint \cref{eq:Xt_psd} as a (nonvarying) semidefinite programming
constraint on $\phi$. This allows us to circumvent the need for time discretization.

In the statement of proposition below, $S^{m}$ (resp.
$\R_d^{m \times m}[t]$) denotes the set of symmetric matrices of size $m$ whose
entries are elements of $\R$ (resp. $\R_d[t]$) for any positive integers $m$
and~$d$.

\begin{proposition}[\cite{tvsdp_2018} Univariate matrix Positivstellensatz]
\label{prop-matrix-positivstellensatz}
Let $m$ and $d$ be positive integers. There exist two (explicit) linear maps
$\lambda_1: S^{\lfloor \frac d2 + 1\rfloor m} \rightarrow \R_d^{m
  \times m}[t]$ and
$\lambda_2: S^{\lfloor \frac d2 + 1\rfloor m} \rightarrow \R_d^{m
  \times m}[t]$ such that for $X \in \R_d^{m \times m}[t]$,
$X(t) \succeq 0 \; \forall t \in[0, T]$ if and only if there exist
positive semidefinite matrices $Q_1$ and $Q_2$ of appropriate sizes that
satisfy the equation
$X = \lambda_1(Q_1) + \lambda_2(Q_2).$
\end{proposition}

\section{Exact Moment Optimization Over piecewise-linear Paths}

% In this section, we restrict present a hierarchy of semidefinite programs that
% is guaranteed to find the best solution to (**) that is piece wise 

\subsection{Search for Piecewise-Linear Paths}

We propose to approximate the shortest path of \ref{eq:opt_shortest_trajectory}
by piecewise-linear paths with a fixed number of pieces. We choose to work
with the family of piecewise linear functions for two reasons. First, they can
uniformly approximate any path over the time interval $[0, T]$ as the number of
pieces grows. Second, fixing a low number pieces often leads to simpler and
smoother paths.

More concretely, we fix a regular subdivision
$\{0, \frac Ts, \frac {2T}s, \ldots, T\}$ of the time interval
$[0, T]$ of size $s$, and we parametrize our candidate
trajectory $\ix(t)$ as follows:
\begin{equation}
\ix(t) = \iu_i + t \iv_i \quad \forall t \in \left[\frac{(i-1)T}s , \frac {iT}s \right), \; \forall i \in [s],\label{eq:piecewise-linear}
\end{equation}
where $\iu_i$, $\iv_i \in \mathbb R^n$ for $i=1,\ldots,s$. We rewrite the
objective function and constraints of \ref{eq:opt_shortest_trajectory} in this
setting directly in terms of $\iu \coloneqq (\iu_1, \ldots, \iu_s)$ and
$\iv \coloneqq (\iv_1, \ldots, \iv_s)$. The objective function in
\ref{eq:opt_shortest_trajectory} can be expressed as
\(\frac{T}s \sum_{i=1}^s \|\iv_i\|,\)
%
% the start and destination constraints become $h_{\text{init}}(\iu, \iv) =
% h_{\text{final}}(\iu, \iv) = 0$, where
% \[h_{\text{init}}(\iu, \iv) \coloneqq \iu_1 - \xinit, h_{\text{final}}(\iu, \iv) \coloneqq \iu_s+T\iv_s - \xfinal,\]
and the obstacle-avoidance constraints become

\[g_{k}(t, \iu_i + t \iv_i) \ge 0 \quad \forall t \in \left[\frac{(i-1)T}s , \frac {iT}s \right), \; \forall i \in [s], k \in [m].\]
To ensure continuity of the path $\ix(t)$ at the grid point~$i\frac{T}s$, for $i=0,\ldots,s$,
we need to impose the additional constraint
$h_i(\iu, \iv) = 0$ with
\[h_i(\iu, \iv) \coloneqq \iu_i + \frac {iT}s \iv_i - \left(\iu_{i+1} + \frac {iT}s \iv_{i+1}\right),\]
and the convention that $\iu_0 = \xinit$, $\iv_0 = 0$, $\iu_{s+1} = \xfinal$,
and $\iv_{s+1}~=~0$.

In conclusion, when specialized to piecewise-linear paths of
type \cref{eq:piecewise-linear}, problem
\ref{eq:opt_shortest_trajectory} becomes
\begin{equation}
  \boxed{
  \tag*{LMP\(\left(s; \D\right)\)}
  \label{eq:opt_shortest_pl_trajectory}
  \begin{aligned}
    &\rho(s; \D) =\min_{\iu_i, \iv_i \in }  \quad \frac{T}s \sum_{i=1}^n \|\iv_i\|\\
    &\textrm{s.t.}\quad  h_i(\iu, \iv) = 0\quad \forall i \in \{0, \ldots, {s+1}\}\\
    %&\iu_i + \frac {iT}s \iv_i = \iu_{i+1} + \frac {iT}s  \iv_{i+1} \quad \forall i \in \{1, \ldots, s\}\\
    &g_k(t, \iu_i + t \iv_i) \ge 0 \;\forall t \in \left[\frac {(i-1)T}s, \frac {iT}s\right],\\
    &\quad\quad\quad\quad\quad\quad\quad\quad\quad\quad\; k \in [m],\; i \in [s].
\end{aligned}
}
\end{equation}

% Recall that the constraints $g_k(t,\iu_i+t\iv_i)\ge 0$, $k=n+1,\ldots,n+m$, state that the trajectory must 
% %stay in some domain $\Omega\subset\R^2$ and 
% avoid some obstacles, while the
% constraints $g_j(t,\iu_i+t\iv_i)=\omega^2-(u_{ij}+t\,v_{ij})^2$, $j=1,\ldots,n$, state that $\ix_i(t)\in\Omega$ for all $t$.

% \vikas{Are equality constraints h introduced earlier dropped above? I agree there is no loss of generality since it can be encoded as $h>=0, h<=0$.}

\ref{eq:opt_shortest_pl_trajectory} is a nonlinear,
nonconvex optimization problem in the variables $(\iu, \iv)$.  The
main difficulty comes from the global constraints
\begin{equation}
\label{constraint-g_k}
g_k(t, \iu_i + t \iv_i) \ge 0\,, \quad \forall t \in \left[\frac {(i-1)T}s, \frac {iT}s\right],\; \forall i\in~  [s].\end{equation}

% In general, the functions $g_k$ appearing in the obstacle avoidance constraints are non-convex. Moreover, we need to impose the constraints for \emph{all} $t$ in the desired interval without resorting to time discretization schemes.

\subsection{A hierarchy of SDPs to find the best piecewise-linear path}
\label{sec:exact_approach}

For any optimal motion problem \ref{eq:opt_shortest_trajectory}, and for any
number of pieces $s$, we present a hierarchy of semidefinite programs
\ref{eq:sdp_pl_trajectory} indexed by a scalar~$r$ with the following
properties: (i)~at every level $r$, the optimal value of
\ref{eq:sdp_pl_trajectory} is a lower bound on that of
\ref{eq:opt_shortest_pl_trajectory}, (ii)~under a compactness assumption, the
the optimal value of \ref{eq:sdp_pl_trajectory} converges monotonically to that
of \ref{eq:opt_shortest_pl_trajectory}, and (iii)~under a compactness and
uniqueness assumption, the optimal solution of
\ref{eq:sdp_pl_trajectory} converges to that of
\ref{eq:opt_shortest_pl_trajectory}.

As a preliminary step, for
each piece $i \in [s]$, we introduce a scalar variable $z_i$ that represents the length
$\|\iv_i\|$ of that piece. Mathematically, we impose the constraints
\begin{equation}
 h^z_i(\iu, \iv, \iz) = 0 \text{ and } z_i \ge 0 \quad i \in
  [s],\label{eq:def_z}  
\end{equation}
where \(h^z_i(\iu, \iv, \iz) =  z_i^2 - \left(\frac{T}s\|\iv_i\|\right)^2\) for $i
\in [s]$.
We introduce the auxiliary variable $z_i$ in this seemingly
complicated way (instead of simply taking $z_i = \|\iv_i\|$) to make
the functions appearing in the objective and constraints of
\ref{eq:opt_shortest_pl_trajectory} polynomial functions.

We are now ready to follow the moment approach presented in
\cref{sec:background-sos}. We fix a positive integer $r$, and we construct the
moment relaxation of order $r$ of problem \ref{eq:opt_shortest_pl_trajectory}.
For that, we need to specify the decision variables, objective, and constraints.
Our decision variable is a truncated sequence $\phi \in \M_r((2n+1) \times s)$
(that should be viewed as a sequence of ``pseudo-moments'' in variables
$\iu \in \R^{n\times s}, \iv\in \R^{n\times s}$, and $ \iz\in \R^{s}$ up to degree $r$).
Intuitively, $\phi$ represents a ``pseudo-distribution'' over candidate
paths. Our objective function is \(\sum_{i=1}^s L_{\phi}(z_i)\),
and our constraints are the moment consistency constraints
  \begin{equation}\label{const:moment-consistency}L_\phi(1)=1 \text{ and } M_\phi(1)\succeq 0,\end{equation}
  the continuity constraints
  \begin{equation}
    \label{const:continuity}L_\phi ((\iu, \iv, \iz)^\ialpha h_i(\iu, \iv) )=0
  \end{equation}
  for all $\ialpha\in \N^{s(2n+1)}_{r-1}$ and
  $i \in \{0, \ldots, s\}$, the obstacle avoidance constraints
  \begin{equation}
    \label{const:obst-avoidance} M_{\phi}(g_k(t, \iu_i + t \iv_i)) \succeq 0\, \; \forall t \in \left[\frac {(i-1)T}s, \frac {iT}s\right]
  \end{equation}
  for all $k \in [m]$ and $i\in [s]$,
  and the constraints
  \begin{equation}
    \label{const:z}L_\phi ((\iu, \iv, \iz)^\ialpha h^z_i(\iu, \iv, \iz))
    = 0 \text{ and } M_\phi(z_i) \succeq 0
  \end{equation}
  coming from the definition of
  $\iz$ in \cref{eq:def_z} for all $\ialpha\in\N^{s(2n+1)}_{r-2}$ and $i\in[s]$.

  In conclusion, the moment relaxation of order $r$ of problem
  \ref{eq:opt_shortest_pl_trajectory} is the SDP
\begin{equation}
  \small
  \boxed{
  \label{eq:sdp_pl_trajectory}
  \tag*{SDP\((r, s; \D)\)}
  \begin{aligned}
    \rho(r, s; \D) = &\min_{\phi \in \PM r {(2n+1) \times s}} \quad  \sum_{i=1}^s L_\phi (z_i)\\
    &\textrm{s.t. $\phi$ satisfies \Cref{const:moment-consistency,const:continuity,const:z,const:obst-avoidance},}
    \end{aligned}
    }
  \end{equation}
  % and we refer to $r$ as the level of the
%
  We emphasize that the objective function of \ref{eq:sdp_pl_trajectory} is
  linear, and that its constraints are valid SDP constraints. Indeed,
  constraints \cref{const:moment-consistency,const:continuity,const:z} are
  (scalar or matrix) linear inequalities, while the time-varying
  inequalities in \eqref{const:obst-avoidance} translates to positive
  semidefinite constraints on the $\phi_\ialpha$'s and some additional auxiliary
  variables in view of Proposition \ref{prop-matrix-positivstellensatz}.

  \Cref{thm:rigorous_sdp_opt_val,thm:rigorous_sdp_opt_sol} below\footnote{The proofs of these results were ommitted to conserve space. They can be found in \cite{githubrepo}} present the
  main results of this section. They are related respectively to the optimal
  value and optimal solution of \ref{eq:sdp_pl_trajectory}.
% In the next theorem, we present a sequence of SDPs indexed by $r \in \N$ whose optimal values are nondecreasing and provide a lower bound on the length of any valid piecewise-linear path of $s$ pieces. We also show under a compactness assumption that these optimal values converge are asymptoically tight.
\begin{theorem}%[Optimal value]
  \label{thm:rigorous_sdp_opt_val}
  Consider the motion planning problem \ref{eq:opt_shortest_trajectory} given by data
  $\D = (\xinit, \xfinal, \{g_{1}, \ldots, g_{m}\})$. The sequence $\{\rho(r,s)\}_{r \in \N}$
  of optimal values of \ref{eq:sdp_pl_trajectory} is
  nondecreasing and is upper bounded by the optimal value $\rho(s; \D)$ of
  \ref{eq:opt_shortest_pl_trajectory}. (In particular, if $\rho(r, s; \D) = \infty$
  for some $r \in \N$, then problem \ref{eq:opt_shortest_pl_trajectory} is
  infeasible.) Furthermore, if
  \begin{equation}
    \label{eq:compactness_assumption}g_{m}(t, \ix) = R^{2} - \|\ix\|^{2} \quad \forall \ix \in \R^{n},\, \forall t \in \R\text{ for some $R > 0$},
  \end{equation}
  then $\rho(r, s; \D) \rightarrow \rho(s; \D)$
  as $r \rightarrow \infty$.
\end{theorem}
Assumption \eqref{eq:compactness_assumption} is needed for technical reasons but
is not restrictive in practive. Indeed, in most motion planning problems, the
configuration space is bounded, in which case we can append the polynomial
$g(t, \ix) \coloneqq R^{2} - \|x\|^{2}$ to the
list of polynomials in $\D$ without loss of generality.

\begin{theorem}%[Optimal solution]
  \label{thm:rigorous_sdp_opt_sol}
  Consider the motion planning problem \ref{eq:opt_shortest_trajectory} given by data
  $\D = (\xinit, \xfinal, \{g_{1}, \ldots, g_{m}\})$. Under assumption
  \eqref{eq:compactness_assumption}, for any $r \in \N$, the optimal value of
  \ref{eq:sdp_pl_trajectory} is attained by some element
  $\phi^r \in \M_r({(2n+1)\times s})$. Furthermore, if
  \ref{eq:opt_shortest_pl_trajectory} has a unique optimal solution
    \begin{equation}
      % \label{eq:unique}
      \ix^*(t)\,:=\,\iu_i^*+t\,\iv_i^*\,,\:t\in
      \left[\frac{(i-1)T}s, \frac{iT}s\right]\quad i\in
      [s],\end{equation}
    then $L_{\phi^r}(\iu_i) \rightarrow \iu_i^*$ and
    $L_{\phi^r}(\iu_i) \rightarrow \iv_i^*$ as $r \rightarrow \infty$
    for $i\in[s]$.
\end{theorem}

\subsection{Detecting optimality of a solution to \ref{eq:sdp_pl_trajectory}}
\input{Imgs/simple_instance_path_planning.tex} The results of
\cref{thm:rigorous_sdp_opt_val,thm:rigorous_sdp_opt_sol} presented in the
previous section are asymptotic. For a given number of pieces $s$ and a given
relaxation order $r$, the optimal value of \ref{eq:sdp_pl_trajectory} provides
only a lower bound on that of \ref{eq:opt_shortest_pl_trajectory}. Recovering
the shortest piecewise-linear path or its corresponding length requires taking
$r$ to infinity in general. The following proposition shows that, if some
conditons that are easily checkable hold, we can get the same recovery
guarantees for finite $r$.

\begin{proposition}
  \label{prop:rank-1} For integers any integers $s$ and $n$, if an optimal
  solution $\phi \in \PM r {(2n+1)\times s}$ of \ref{eq:sdp_pl_trajectory}
  satisfies
  \(L_{\phi}(\|\iu_{i}\|^r) = \|L_{\phi}(\iu_i)\|^r, L_{\phi}(\|\iv_{i}\|^r) = \|L_{\phi}(\iv_i)\|^r\),
  and \(L_{\phi}(z_{i}^{r}) = L_{\phi}(z_{i})^{r}\) %\label{eq:rank_one_cdt}
  for $i\in[s]$, then the piecewise-linear path
    \begin{equation*}
      % \label{eq:unique}
      \ix^*(t)\,:=\,L_{\phi}(\iu_i)+t\,L_{\phi}(\iv_i)\quad \forall t\in
      \left[\frac{(i-1)T}s, \frac{iT}s\right],\;\forall i\in
      [s],\end{equation*}
    is optimal for \ref{eq:opt_shortest_pl_trajectory}.
  \end{proposition}
Other than its obvious practical benefit, the result of \cref{prop:rank-1} inspires the iterative approach we present in \cref{sec:heuristic}.

\begin{example}
  \label{exp:simple_instance}
Consider the simple instance of \ref{eq:opt_shortest_trajectory} in dimension $n=2$ given by data $\D = (\xinit, \xfinal, \{g_{1} \ldots, g_{5}\})$, where
\(\xinit =
  (0, -1)
  ^T\),
  \(\xfinal =
(0, 1)^T\),
  \(g_1(t, \ix) = 1 - x_1\), \(g_3(t, \ix) = 1 + x_1,\)
  \(g_3(t, \ix) = 1 - x_2\), \(g_4(t, \ix) = 1 + x_2,\)
  %
  % \(g_5(t, \ix) = x_1^2+(x_2+\frac12 (1 - t))^2 - \left(\frac6{10}\right)^{2}.\)
  \(g_{5}(t, \ix) = (\ix_{1}+\frac{1}{3})^2 + (\ix_{2}-\frac15)^2 - t (\ix_{1}+\frac{1}{3})^3 - (\frac{1}{2})^2.\)
(See \cref{fig:simple_instance_path_planning} for a plot of this setup.)
We search for paths that are piecewise-linear of the form in
\eqref{eq:piecewise-linear} with $s=2$ pieces. By computing the optimal values
of \ref{eq:sdp_pl_trajectory} for $r \in \{3, 4, 5, 6\}$, we obtain the
nondecreasing sequence of lower bounds
  \begin{center}
    \begin{tabular}{l|r|r|r|r|}
      $r$ & 3 & 4 & 5 & 6\\
      \hline
      \ref{eq:sdp_pl_trajectory}& 0.75 & 1.81 & 2.09 & 2.14\\
    \end{tabular}
    \vspace{.2cm}
  \end{center}
  on the length of any piecewise-linear path with $2$ pieces that starts in
  $\xinit$, ends at $\xfinal$, and avoids the obstacles given by the polynomials
  $\{g_1, \ldots, g_{5}\}$. In particular, no such path has length smaller than
  $2.14$. We check numerically that for $r=6$, the optimal solution $\phi$ of
  \ref{eq:sdp_pl_trajectory} returned by the solver
  satisfies the requirement of~\cref{prop:rank-1}, and we extract from $\phi$
  the path plotted in \cref{fig:simple_instance_path_planning} whose length is~$2.14$.
\end{example}

\subsection{A sparse version of the SDP hierarchy
  \ref{eq:sdp_pl_trajectory}}
\label{sec:sparse}
% TODO: make remark that the sparsity can be pushed further since
% $z_i$ only appears in the cost criterion and in the equality
% constraints $z_i^2=\Vert\iv_i\Vert^2$.

In this section we briefly describe how one may reduce the size of the
semidefinite programs \ref{eq:sdp_pl_trajectory} by exploiting an inherent
sparsity of \ref{eq:opt_shortest_pl_trajectory}. If we partition the decision
variables $(\iu, \iv, \iz)$ of \ref{eq:opt_shortest_pl_trajectory} as
\(V_1 \cup \cdots \cup V_s,\) where for each $i \in [s]$,
$V_{i} \coloneqq \{ \iu_i, \iv_i, z_i, \iu_{i+1}, \iv_{i+1}, z_{i+1}\}$, then
each constraint that appear in \cref{const:continuity},
\cref{const:obst-avoidance}, or \cref{const:z} involves only the variables of
exactly one of the $V_i$'s. Furthermore, the family $\{V_1, \ldots,V_s\}$
satisfies the \emph{Running Intersection Property} (RIP), that is,
\begin{equation}\tag{RIP}
\label{eq:RIP}
\forall i\in [s-1], \;\exists k \leq i, \left(V_1\cup \ldots \cup V_{i}\right) \cap V_{i+1} \subset V_{k}.
\end{equation}

Following \cite{Waki_2006,Lasserre_2006}, we replace the single truncated
sequence of ``pseudo-moments'' $\phi \in \PM r {(2n+1)\times s}$ in all
variables of $V$ with $s$ truncated sequences
$\phi_{1}, \ldots, \phi_{s} \in \PM r {2n+1}$, where for each $i \in [s]$,
$\phi_{i}$ is a truncated sequence of ``pseudo-moments'' in the variables of
$V_{i}$. Intuitively, $\phi_{i}$ represents a ``pseudo-distribution'' from which
the $i\text{-th}$ piece of our candidate piecewise-linear path is sampled.
Without entering into details beyond the scope of this paper we can prove that
\cref{thm:rigorous_sdp_opt_val,thm:rigorous_sdp_opt_sol} hold if
\ref{eq:sdp_pl_trajectory} is replaced with the SDP

\hspace{-1cm}
\begin{equation}
  \small
  %\hspace{-1cm}
  \label{eq:sdp-sparse}
  \tag*{SparseSDP($r,s; \D$)}
  \hspace{-1cm}
  \boxed{
  \begin{aligned}
    &\rho'(r, s; \D) =\min_{\phi_i} \quad  \sum_{i=1}^s L_{\phi_i} (z_i)\\
    &\textrm{s.t.} \quad  L_{\phi_i}(1)=1, \; M_{\phi_i}(1)\succeq0,& i\in [s-1]\\\
    % \quad &M_{r-1,\phi_i}(\theta(u_{ij}))\,,M_{r-1,\phi_i}(\theta(u_{i+1j}))\,\succeq0\,,
    % \quad j\in [n]\,,\:i\in [s-1]\\
    % \quad &M_{r-1,\phi_i}(\theta(v_{ij}))\,,M_{r-1,\phi_i}(\theta(v_{i+1j}))\,\succeq0\,,
    % \quad j\in [n]\,,\:i\in [s-1]\\
    & L_{\phi_i}  \left((V_i, V_{i+1})^\ialpha \tilde h_i(V_i, V_{i+1})\right)=0\,&
    \forall \ialpha\in \N^{4n}_{2r-1}, \; i \in [s],\\
    & L_{\phi_i}  \left((V_i, V_{i+1})^\ialpha \tilde h^z_i(V_i)\right)=0\,&
    \forall \ialpha\in \N^{4n}_{2r-2}, \; i \in [s],\\
    & M_{\phi_i}(g_k(t, \iu_i + t \iv_i)) \succeq  0 &\forall t \in \left[\frac {(i-1)T}s,
      \frac {iT}s\right],\\
    & &k \in [m]\,;\, i \in [s-1],
    \end{aligned}
}
\end{equation}
where for each $i \in \{0, \ldots, s\}$, $\tilde h_i$ is the polynomial function
such that $\tilde h_i(V_i, V_{i+1}) = h_i(\iu, \iv)$, and for each $i \in [s]$,
$\tilde h_i^z$ is the polynomial function such that
$\tilde h^z_i(V_i) = h^z_i(\iu, \iv, \iz)$. The main feature of
\ref{eq:sdp-sparse} when compared to \ref{eq:sdp_pl_trajectory} is that its
constraints involve localizing matrices with pseudo-moments on
$2(2n+1)$ variables (instead of $s(2n+1)$ variables in
\ref{eq:sdp_pl_trajectory}). For more details about the use of sparsity in
polynomial optimization problems, interested reader is referred to
\cite{lasserre2010moments}.

\input{Imgs/setup_2d_moving_obstacles.tex}
% \begin{filecontents*}{comparisontable.csv}
% speedobs,n,SOS,RRT,RRT_tv,NLP
% 0.0,2,0.6,0.4,0.5,0.0
% 0.0,3,1.0,0.5,0.4,0.7
% 0.0,4,1.0,0.6,0.0,0.8
% 1.0,2,0.5,0.3,0.5,0.2
% 1.0,3,1.0,0.4,0.4,0.6
% 1.0,4,1.0,0.7,0.0,0.9
% \end{filecontents*}

% \begin{table}[htbp] \centering
%   \begin{tabular}{l|l|l|l|l}%

%   \bfseries Obstacles moving? & \bfseries Dimension & \bfseries Heuristic & \bfseries RRT & \bfseries NLP
%     \csvreader[head to column names]{comparisontable.csv}{}{\\\hline\speedobs & \n & \SOS & \RRT & \NLP}
%     \end{tabular}
%   % \csvautotabular{comparisontable.csv}
%   \caption{Percentage of time a given methods succeeds in finding a valid path}
%   \label{tab:SOS-RRT-NLP}
% \end{table}
\newcommand\formattedtable[1]{%
  \begin{tabular}{l|l|l|l|l|l|l|l|l|l|l|l|l|l|l|l|l|l|l|l|l|l|l|l|l|l|l|l|l|l|l|}
    & \multicolumn{10}{|l|}{$n=2$} &
      \multicolumn{10}{|l|}{$n=3$}\\
    % \multicolumn{10}{|l|}{$n=4$}\\\hline
     #1
\end{tabular}%
}
\newcommand\HL[1]{{\bf #1}}
\begin{table*}[!h]
  \vspace{-.5cm}
\centering
\begin{tabular}{llrrrr|rrrr}
\toprule
\multirow{3}{*}{$n$} & \multirow{2}{*}{Methods} & \multicolumn{4}{c}{Static Obstacles} & \multicolumn{4}{c}{Dynamic Obstacles}\\
\cmidrule{3-6} \cmidrule{7-10} \\
{} & {} & success rate& length & smoothness & solve time & success rate& length & smoothness & solve time\\
  \midrule
\multirow{3}{*}{$2$} &RRT &40\% &3.73&0.06&0.02{} &50\% &3.59&0.12&0.06\\
{} &NLP &0\% &nan&nan&nan{} &20\% &3.4&0.06&0.06\\
{} &\HL{MMP} &\HL{60}\% &\HL{3.0}&\HL{0.03}&{0.43} &\HL{50} \% &\HL{2.85}&\HL{0.03}&{0.43}\\
\hline
\multirow{3}{*}{$3$} &RRT &50\% &5.74&0.13&0.12{} &40\% &4.82&0.1&0.23\\
{} &NLP &70\% &3.44&0.05&0.12{} &60\% &3.48&0.06&0.17\\
{} &\HL{MMP} &\HL{100}\% &{3.5}&\HL{0.04}&{0.47}{} &\HL{100}\% &{3.55}&\HL{0.04}&{0.47}\\
\hline
\multirow{3}{*}{$4$} &RRT &60\% &7.67&0.15&1.43{} &0\% &nan&nan&nan\\
{} &NLP &80\% &3.99&0.08&0.25{} &90\% &4.34&0.1&0.2\\
{} &\HL{MMP} &\HL{100}\% &{4.11}&\HL{0.05}&{0.55}{} &\HL{100}\% &\HL{4.1}&\HL{0.05}&{0.55}\\
  \bottomrule
\end{tabular}
\caption{\label{tbl:comparison_sos_rrt_nlp}\small Average success, smoothness, and solve-time comparison of RRT, NLP and MMP (proposed) methods over 10 static and dynamic motion planning problems.
\vspace{-.5cm}
}
\end{table*}

%%% Local Variables:
%%% mode: latex
%%% TeX-master: "../Path_Planning_Using_Moments"
%%% End:

% no subfigure caption label.
\captionsetup[subfigure]{labelformat=empty}
\newcommand\pendulumframes[1]{%
  \hspace{-1.5cm}
  \foreach \i in {0,1, ..., 9}{
    \frame{\subfloat[$t=\frac{\i}{10}$]{\includegraphics[width=.09\linewidth, trim= 5.cm 5cm 7cm 3cm, clip]{Imgs/double_pendulum/double_pendulum_0.\i_#1.png}}}~}
    \frame{\subfloat[$t=\frac{10}{10}$]{\includegraphics[width=.09\linewidth, trim= 5.cm 5cm 7cm 3cm, clip]{Imgs/double_pendulum/double_pendulum_1.0_#1.png}}}~}

 \newcommand\pendlumsetup[7] {%
   \begin{tikzpicture}[scale=#6]%[background rectangle/.style={fill=red!100}, show background rectangle]
     % boudning box
     \path (-3.7cm,0cm) rectangle (1cm,2.5cm);
    % save length of g-vector and theta to macros
    \pgfmathsetmacro{\scale}{3}
    \pgfmathsetmacro{\larm}{.48 * \scale}
    \pgfmathsetmacro{\ldashed}{.5 * \scale}
    \pgfmathsetmacro{\atheta}{#1 * 180/3.14}
    \pgfmathsetmacro{\aalpha}{#2 * 180/3.14}
    \pgfmathsetmacro{\agamma}{#3 * 180/3.14}

    \pgfmathsetmacro{\bx}{-1 * \scale}
    \pgfmathsetmacro{\by}{0}
    \pgfmathsetmacro{\cx}{.14 * \scale}
    \pgfmathsetmacro{\cy}{.35 * \scale}
    \coordinate (centr1) at (\bx, \by);
    \coordinate (centr2) at (\cx, \cy);

    \draw[thick, blue] (centr1) -- ++(0+\atheta:\larm) coordinate (boba);
    \draw[thick, blue] (boba) -- ++(\aalpha:\larm) coordinate (bobb);

    \draw[thick,black!20!green] (centr2) -- ++(0+\agamma:\larm) coordinate (bobc);

    \ifnum #5 > 0
    \draw[dashed,gray,-] (centr1) -- ++ (\ldashed,0) node (marya) [black,below]{$ $};
    \draw[dashed,gray,-] (boba) -- ++ (\ldashed,0) node (maryb) [black,below]{$ $};
    \draw[dashed,gray,-] (centr2) -- ++ (\ldashed,0) node (maryc) [black,below]{$ $};
    \pic [draw, ->, "$\theta$", angle eccentricity=1.5] {angle = marya--centr1--boba};
    \pic [draw, ->, "$\alpha$", angle eccentricity=1.5] {angle = maryb--boba--bobb};
    \pic [draw, ->, "$\gamma$", angle eccentricity=1.5] {angle = maryc--centr2--bobc};
    \fi
    \filldraw [fill=blue] (centr1) circle[radius=0.1];
    \filldraw [fill=blue] (boba) circle[radius=0.1];
    \filldraw [fill=blue] (bobb) circle[radius=0.1];
    \filldraw [fill=black!30!green] (bobc) circle[radius=0.1];
    \filldraw [fill=black!30!green] (centr2) circle[radius=0.1];
    \ifnum #7 > 0
    \node at (-1.1, -1) { #4};
    \else
    \node at (-1.5, -1.5) { #4};
    \fi
\end{tikzpicture}}

\newcommand\exptitle[1]{%
\begin{tikzpicture}
  \path (-1, -1) rectangle (1, 1);
  \node at (0, 0) {\Large #1};
\end{tikzpicture}
}
\begin{figure*}[!ht]
  \centering
  \pendlumsetup{1}{1}{4.71}{Initial configuration}{1}{.7}{1}\quad \quad \quad
  \pendlumsetup{0}{0}{2.71}{Goal configuration}{0}{.7}{1}

\begin{filecontents*}{Imgs/double_pendulum/opttrajectoryall_small.csv}
0.6406926634785213,0.9738640751679185,4.60249705161545,1.522080668540012,0.43957708576612287,4.3356384926197995,0.9857908561106178,0.7175208139118857,4.216168340840956
0.2414622895657671,0.9448241586878279,4.480394908538518,2.102170300251136,-0.18311504116040733,3.9170268395476997,0.9700029184557486,0.40365505159175863,3.664812074681252
-0.15776808434698664,0.9157842422077374,4.358292765461585,1.9194623879508868,-0.056412723143772756,3.621451809121313,0.9542149808008794,0.08978928927163188,3.113455808521548
-0.4501163062684148,0.8360304960343219,4.182468568687676,1.6519991929827067,0.1535556443109914,3.33954751454445,0.9472486939361842,-0.04621311684906841,2.874513391879674
-0.6607311178435347,0.7174955859777756,3.9655628014396056,1.1009584776973225,-0.1296886732878051,2.7323459866902144,0.9470283753227984,-0.046202368229030386,2.8744756836926117
-0.8713459294186544,0.5989606759212294,3.7486570341915364,0.5184091490433598,-0.4677344003369972,2.0890003218051607,0.9468080567094126,-0.04619161960899237,2.8744379755055496
-0.9605200807955784,0.47251803195430614,3.5214644468131215,0.25308481889787965,-0.2758464351730544,2.3341412065972698,0.8803744055684097,-0.04295054247027579,2.863067659909714
-0.6506748934421658,0.32009286035614276,3.260472360868466,0.02300771080889816,-0.025076948652095865,2.678002819131288,0.5963826618366644,-0.02909552877018682,2.8144616342564803
-0.3408297060887536,0.1676676887579796,2.9994802749238105,0.0,0.0,2.7123889803846897,0.3123909181049196,-0.015240515070097861,2.7658556086032466
-0.03098451873534147,0.015242517159816449,2.7384881889791552,0.0,0.0,2.7123889803846897,0.02839917437317474,-0.0013855013700089097,2.717249582950013
\end{filecontents*}
\DTLloaddb[noheader]{traj}{Imgs/double_pendulum/opttrajectoryall_small.csv}

\hspace{-0.5cm}
\vspace{-.2cm}
\exptitle{RRT}
\DTLforeach{traj}{\x=Column4,\y=Column5,\z=Column6}{%
  \boxed{
    \pendlumsetup{\x}{\y}{\z}{$t = \frac{\arabic{DTLrowi}}{10}$}{0}{.3}{0}~\hspace{-.3cm}%
  }
}
\vspace{-.2cm}
\hspace{-0.5cm}
\exptitle{NLP}
\DTLforeach{traj}{\x=Column7,\y=Column8,\z=Column9}{%
  \boxed{
    \pendlumsetup{\x}{\y}{\z}{$t = \frac{\arabic{DTLrowi}}{10}$}{0}{.3}{0}~\hspace{-.3cm}%
  }
}

\hspace{-0.5cm}
\vspace{-.2cm}
\exptitle{MMP}
\DTLforeach{traj}{\x=Column1,\y=Column2,\z=Column3}{%
  \boxed{
    \pendlumsetup{\x}{\y}{\z}{$t = \frac{\arabic{DTLrowi}}{10}$}{0}{.3}{0}~\hspace{-.3cm}%
  }
}

% \DTLforeach{traj}{\x=Column4,\y=Column5,\z=Column6}{%
% \pendlumsetup{\x}{\y}{\z}{}{0}%{0}
% }

% \DTLforeach{traj}{\x=Column7,\y=Column8,\z=Column9}{%
% \pendlumsetup{\x}{\y}{\z}{}{0}%{\small Nonlinear Programming Baseline}{0}
% }
%   % {\bf RRT}

  % \pendulumframes{rrt}

  % {\bf NLP}

  % \pendulumframes{NLP}

  % {\bf MMP}
 
  % \pendulumframes{sos}

\caption{\label{fig:bimanual}Performance comparison of RRT, NLP, and MMP (proposed) methods on a bimanual planar manipulation task.
\vspace{-.5cm}}
\end{figure*}

%%% Local Variables:
%%% mode: latex
%%% TeX-master: "../Path_Planning_Using_Moments"
%%% End:

\section{Moment Motion Planner: An Iterative Optimization Procedure Over piecewise-linear Paths}
\label{sec:heuristic}

\begin{algorithm}
\caption{MMP: Moment Motion Planner}

\begin{algorithmic}[1]

%\Procedure{Roy}{$a,b$}       \Comment{This is a test}
  \State {\bf Input:} Data $\D = (\{\xinit, \xfinal, \{g_{1}, \ldots, g_{m}\})$
  % (where $\xinit \in $ is the starting configuration, $\xfinal\in $ is the final configuration, and
  %   $g_1, \ldots, g_k \in \R[t, \ix]$ are polynomial functions describing
  %   the obstacles),
  order $r$ of the moment relaxation, number of
    iterations $N$, trade-off constant $\lambda > 0$.

    \State Initialize $\phi^{(0)} = (\phi^{(0)}_{1}, \ldots, \phi^{(0)}_{s})$
    randomly, where for each $i \in [s]$, $\phi^{(0)}_{i} \in \PM r {2n}$.
    \For{$t = 1, \ldots, N$} %\Comment{put some comments here}
    \State Let $\phi^{(t)}$ be a minimizer of~(\ref{eq:lin-heuristic-obj}) with
    $\bar \phi = \phi^{(t)}$ subject to \cref{eq:heuristic-consistency},
    \cref{eq:heuristic-cont}, and \cref{eq:heuristic-obs}.
    \EndFor \label{for:heuristic_loop}
    \State Return the piecewise-linear path defined by
    \[\ix^*(t)\,:=\,L_{\phi_{i}^{(N)}}(\iu_i)+t\,L_{\phi_{i}^{(N)}}(\iv_i)\]
    for every $t\in \left[\frac{(i-1)T}s, \frac{iT}s\right]$ and every
    $i\in [s].$
\end{algorithmic}
\end{algorithm}

% For a given motion planning problem, solving \ref{eq:sdp_pl_trajectory} at some
% order $r$ provides only a lower bound on the minimum length of all valid piecewise
% linear path with $s$ pieces.   in the previous section is that it only gives lower
% bounds

% scalability. Indeed,
% A downside of the fact

As we have seen in the previous section, the optimal values $\rho(r, s)$ of the
SDPs in the hieararchy \ref{eq:sdp_pl_trajectory} are nondecreasing lower bounds
on the optimal value $\rho(s)$ of \ref{eq:opt_shortest_trajectory}. As a
downside, a feasible path cannot possibly be extracted from a solution of one
of these SDPs (at order, say, $r$) unless $r$ is large enough so that
$\rho(r, s) = \rho(s)$. The order~$r$ needed for that to happen is in general
prohibitively large.

To adress this issue, we present in \cref{for:heuristic_loop} a more practical
motion planner called MMP. MMP is also based on a moment relaxation, but has two
distinctive features when compared to \ref{eq:sdp_pl_trajectory}: (i) it
produces feasible paths already for low orders $r$ (taking $r=2$ produced good
results in all of our benchmarks) and (ii) the optimal values produced by MMP
are not necessarily lower bounds on $\rho(s)$. In other terms, MMP
trades off some the theoretical guarantees of \ref{eq:sdp_pl_trajectory} for
more efficiency.

MMP is an iterative algorithm. At every iteration, we solve an SDP that is
similar in spirit to \ref{eq:sdp_pl_trajectory} with a few key differences.
First, we drastically decrease the number of decision variables. We completely
discard the variable $\iz$, and we take inspiration from the sparsity
considerations reviewed in \cref{sec:sparse} to partition the remaining
variables $(\iu, \iv)$ of \ref{eq:opt_shortest_pl_trajectory} as
\(W_1 \cup \cdots \cup W_s,\) where for each $i \in [s]$,
$W_{i} \coloneqq \{ \iu_i, \iv_i\}$. Then, we take as decision variables of our
inner SDP $s$ truncated sequences $\phi \coloneqq (\phi_{1}, \ldots, \phi_{s})$,
where for each $i \in [s]$, $\phi_{i}\in \PM r {2n}$ is a truncated sequence of
``pseudo-moments'' in variables $W_{i}$. Intuitively, each $\phi_{i}$ represents
a ``pseudo-distribution'' from which the $i\text{-th}$ piece of our candidate
piecewise-linear path is sampled. Note that the family of sets
$\{W_1, \ldots W_{s}\}$ does not satisfy the \eqref{eq:RIP} property anymore.
This is the main reason why MMP lacks some of the theoretical guarantees of the
moment relaxation \ref{eq:sdp_pl_trajectory}.

Then, we adapt the constraints of our inner SDP to our new choice of decision
variables. In addition to the classical moment-consistency constraints
\begin{equation}
  \label{eq:heuristic-consistency}
  L_{{\phi_{i}}}(1) = 1 \text{ and  } \M_{{\phi_{i}}}(1) \succeq 0 \; \forall i \in [s],
\end{equation}
we impose the the continuity constraints
  % TODO: Change to (8)
  \begin{equation}\label{eq:heuristic-cont}
    L_{\phi_i} ((\iu_i + \frac {iT}s \iv_i)^\ialpha) = L_{\phi_{i+1}}((\iu_{i+1} + \frac {iT}s \iv_{i+1})^\ialpha ) \; \forall \ialpha \in \N_{r-2}^{2n}
  \end{equation}
  between endpoints of pieces $i$ and $i+1$ for each $i \in [s]$,
  % Replace by 9
  and the obstacle-avoidance constraints
  \begin{equation}\label{eq:heuristic-obs}
    M_{\phi_i} (g_k(t, \iu_{i} + t \iv_{i}) ) \succeq  0 \, \forall t \in \left[\frac {(i-1)T}s, \frac {iT}s\right]
  \end{equation}
  for each obstacle $k \in [m]$ and for each piece $i \in [s]$.

  Finally, let us explain our choice of objective function. Motivated by
  \cref{prop:rank-1}, we would ideally like to take the objective function of
  our SDP to be
  \(\sum_{i=1}^{s} \|L_{{\phi_{i}}}(\iv_{i})\| + \lambda J(\phi), \)
  where $\lambda > 0$ and
  \begin{equation}
    \begin{split}
      J(\phi) =  &\sum_{i=1}^{s} L_{{\phi_{i}}}(\|\iu_{i}\|^{r}) - \|L_{{\phi_{i}}}(\iu_{i})\|^{r} \\&+ L_{{\phi_{i}}}(\|\iv_{i}\|^{r}) - \|L_{{\phi_{i}}}(\iv_{i})\|^{r}
  \end{split}
    \label{eq:heuristic-obj}
\end{equation}
The intuition is that, if $J(\phi) = 0$ for some $\phi = (\phi_{1}, \ldots, \phi_{s})$
satisfying \cref{eq:heuristic-cont,eq:heuristic-obs}, then the path
\[\ix^*(t)\,:=\,L_{\phi_{i}}(\iu_i)+t\,L_{\phi_{i}}(\iv_i)\;\forall t\in
      \left[\frac{(i-1)T}s, \frac{iT}s\right],\;\forall i\in
      [s],\]
is feasible to \ref{eq:opt_shortest_trajectory}. The constant $\lambda$ controls the
trade-off between minimizing the length of the path and enforcing that the path
is feasible. The issue with objective function \eqref{eq:heuristic-obj} is that the
function $J$ is nonconvex. As a workaroud, we replace $J$ in
\eqref{eq:heuristic-obj} with its linearization around a reference point
$\bar \phi = (\bar \phi_{1}, \ldots, \bar \phi_{s})$, leading to the objective
function
  \begin{equation}
  \sum_{i=1}^{s} \|L_{{\phi_{i}}}(\iv_{i})\| + \lambda \bar J(\phi; \bar \phi),
    \label{eq:lin-heuristic-obj}
\end{equation}
  where $\bar J(\phi; \bar \phi)$ is given by
  \begin{equation*}
    \begin{split}  J(\bar \phi) &+ \sum_{i=1}^{s}
    L_{{\phi_{i}}}(\|\iu_{i}\|^{r}) - (r-1) L_{{\phi_{i}}}(\iu_{i}) \|L_{{\bar \phi_{i}}}(\iu_{i})\|^{r-1} \\&+
    L_{{\phi_{i}}}(\|\iv_{i}\|^{r}) - (r-1) L_{{\phi_{i}}}(\iv_{i}) \|L_{{\bar \phi_{i}}}(\iv_{i})\|^{r-1}
   \end{split}
 \end{equation*}

 In the $t\text{-th}$ iteration of our iterative approach, we take the optimal
 solution $\phi^{{(t-1)}}$ obtained from solving the inner SDP at iteration
 $t-1$, and uses that as $\bar \phi$. The elements of $\phi^{(0)}$ are initialized from some random distribution. Gaussian initializaton seems to work well in practice.
Note that the overall time complexity of  \cref{for:heuristic_loop} is polynomial in the dimension $n$, the number of iterations $N$, and the number of pieces $s$.

\section{Numerical Results}

For animations of motion planning problems and code to reproduce numerical results, please see \cite{githubrepo}. % \\ \url{https://github.com/bachirelkhadir/PathPlanningSOS.jl}

\subsection{MMP vs NLP vs RRT} {\bf Setup.} In each dimension
$n \in \{2, \ldots, 4\}$, we generate $10$ motion planning problems where the
path is constrained to live in the unit box $B = [-1, 1]^{n}$ and must avoid
$10$ static or dynamic spherical obstacles. More precisely, each motion planning
problem is given by data $\D = (\xinit, \xfinal, \{g_1, \ldots, g_m\})$, where
$\xinit = (-1, \ldots, -1) \in \R^{n}, \xfinal = (1, \ldots, 1) \in \R^{n}$,
$i \in [n]$, $g_{i}(\ix) = 1 - x_{i}$, $g_{i+n}(t, \ix) = 1 + x_{i}$ for
$i \in [n]$,
$g_{2n+k}(t, \ix) = \|\ix - (\ic_{k} + t \iv_{k})\|^{2} - \left(\frac{2}{10}\right)^2$.
The centers $\ic_{k}$ are sampled uniformly at random from $B$, the velocities
$\iv_{k}$ are either identically zero in the static case, or sampled uniformly
from $B$ in the dynamic case. See \cref{fig:setup_3d_moving_obs} for an example
of this setup in dimension $n = 2$.

\textbf{Comparison.} In \cref{tbl:comparison_sos_rrt_nlp}, we compare our MMP
solver (with $r=2$, $N=20$, and $\lambda=0.1$) against a classical sampling-based
technique (RRT) and basic nonlinear programming baseline (NLP) implemented with the
help of the KNITRO.jl package~\cite{knitro}. MMP consistently achieves higher
success rates, significantly shorter and smoother trajectories (the smoothness of a
path $\ix(t)$ is given by
$\int_{0}^{T} \left(\dot \ix(t) - \int_{0}^{T} \dot \ix(s) {\rm d}s\right)^{2} {\rm d}t$).
The solve times are higher but remain highly practical.

\begin{figure}[!h]
\begin{minipage}{.05\textwidth}
\tdplotsetmaincoords{70}{50}
\begin{tikzpicture}[tdplot_main_coords]
\draw[->] (0,0,0) -- (2,0,0) node[right] {$\theta$};
\draw[->] (0,0,0) -- (0,2,0) node[right] {$\alpha$};
\draw[->] (0,0,0) -- (0,0,2) node[above] {$\gamma$};
\end{tikzpicture}
\end{minipage}\quad\quad\quad\quad~
  \begin{minipage}{.16\textwidth}
  \includegraphics[width=.8\textwidth]{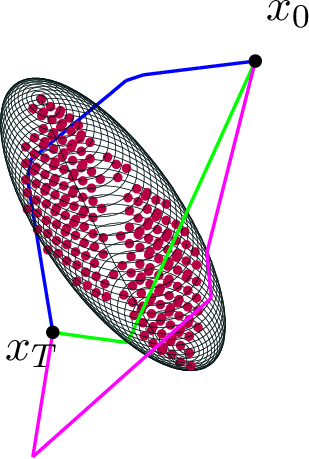}
\end{minipage}~
\begin{minipage}{.1\textwidth}
  % Dummy figure to get the legend
 \begin{tikzpicture}
    \begin{axis}[%
    hide axis,
    xmin=10,
    xmax=50,
    ymin=0,
    ymax=0.4,
    legend style={draw=white!15!black,legend cell align=left}
    ]
    \addlegendimage{ultra thick, blue}
    \addlegendentry{MMP};
    \addlegendimage{ultra thick,magenta}
    \addlegendentry{RRT};
    \addlegendimage{ultra thick, black!30!green}
    \addlegendentry{NLP}
    \end{axis}
  \end{tikzpicture}
\end{minipage}
\caption{\label{fig:obs-joint-space}\small Plot of the paths found by RRT, NLP, and MMP (proposed) for the bimanual planning manipulation task of \cref{fig:bimanual}. The red dots depict values of the joint angles $(\alpha, \beta, \gamma)$ that would make the two arms collide. The black wireframe depicts the smallest enclosing ellipsoid containing the red dots.}
\vspace{-.5cm}
\end{figure}

%%% Local Variables:
%%% mode: latex
%%% TeX-master: "../Path_Planning_Using_Moments"
%%% End:

\subsection{Bimanual Manipulation} As a proof of concept, we also consider a
bimanual manipulation task requiring two two-link arms working collaboratively
to go from an initial to goal configuration without colliding (see
\cref{fig:bimanual}). For visualization, we restrict attention to planar
manipulation problems involving planning in a configuration space of $3$ joint
angles. First, we lift the obstacle set in cartesian space to joint angle space
by evaluating all collision configurations on a grid over the $3$ joint angles.
We then fit an enclosing ellipsoid (see \cref{fig:obs-joint-space}) to obtain a
semialgebraic description of the obstacle region in configuraton space. RRT
finds a feasible path, but produces a jerky path involving an unnecessary
recoiling of one of the arms. NLP fails to find a feasible path at all. MMP, with
$3$ segments, succeeds in finding the shortest and smoothest path.

\section*{Acknowledgements} We thank Ed Schmerling and Brian Ichter for references on motion planning with time-varying obstacles, Carolina Parada and Jie Tan for robot mobility research frameworks, and Amirali Ahmadi for guidance on SoS programming.

\bibliographystyle{plain}
\bibliography{citations}

\end{document}